\documentclass{article}

\usepackage[final]{corl_2018} 

\usepackage[utf8]{inputenc} 
\usepackage[T1]{fontenc}    
\usepackage{hyperref}       
\usepackage{url}            
\usepackage{booktabs}       
\usepackage{amsfonts}       
\usepackage{nicefrac}       
\usepackage{microtype}      
\usepackage{graphicx}       
\usepackage{amsmath}        
\usepackage{mymacros}
\usepackage{lmodern}        
\usepackage{makecell}       

\title{Self-Supervised Sparse-to-Dense: Self-Supervised Depth Completion from \lidar and Monocular Camera}

\author{
  Fangchang Ma, Guilherme Venturelli Cavalheiro, Sertac Karaman\thanks{The authors are affiliated with the Department of Aeronautics and Astronautics (AeroAstro), and the Laboratory for Information \& Decision Systems (LIDS), both at MIT.} \\
  Massachusetts Institute of Technology\\
  \texttt{\{fcma,guivenca,sertac\}@mit.edu}
}

\begin{document}

\maketitle


\begin{abstract}
Depth completion, the technique of estimating a dense depth image from sparse depth measurements, has a variety of applications in robotics and autonomous driving. However, depth completion faces 3 main challenges: the irregularly spaced pattern in the sparse depth input, the difficulty in handling multiple sensor modalities (when color images are available), as well as the lack of dense, pixel-level ground truth depth labels. 
In this work, we address all these challenges. Specifically, we develop a deep regression model to learn a direct mapping from sparse depth (and color images) to dense depth. We also propose a self-supervised training framework that requires only sequences of color and sparse depth images, without the need for dense depth labels. Our experiments demonstrate that our network, when trained with semi-dense annotations, attains state-of-the-art accuracy and is the winning approach on the KITTI depth completion benchmark\footnote{\url{http://www.cvlibs.net/datasets/kitti/eval_depth.php?benchmark=depth_completion}} at the time of submission. Furthermore, the self-supervised framework outperforms a number of existing solutions trained with semi-dense annotations.
\end{abstract}


\section{Introduction}\label{sec:introduction}

\begin{figure}[htbp]
\centering
\begin{minipage}{1.0\linewidth}
\newcommand{\figWidth}{ 0.5\linewidth } 
\setlength\tabcolsep{0.4mm} 
\begin{tabular}{ c c }
  \begin{minipage}[m]{\figWidth}\centering
  \includegraphics[width=\linewidth]{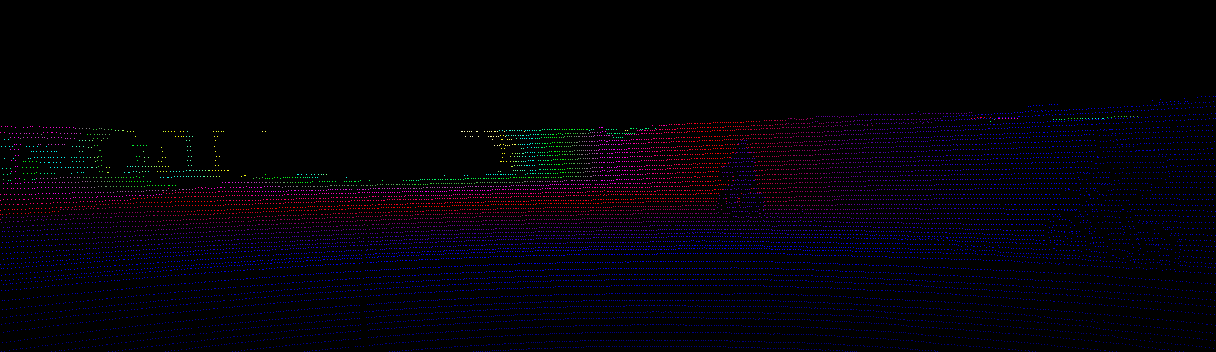} \\
  (a) raw \lidar scans
  \end{minipage}
  & 
  \begin{minipage}[m]{\figWidth}\centering
  \includegraphics[width=\linewidth]{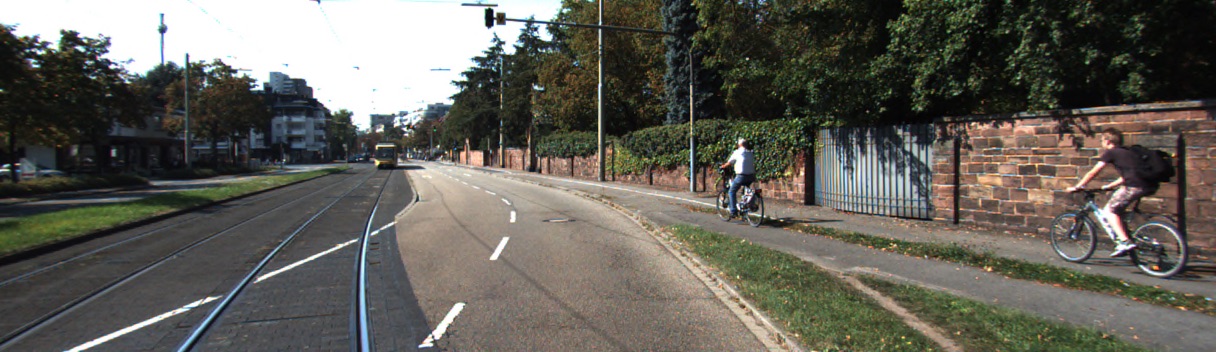} \\
  (b) RGB
  \end{minipage} \\ 
  \vspace{-1em} \\
  \begin{minipage}[m]{\figWidth}\centering
  \includegraphics[width=\linewidth]{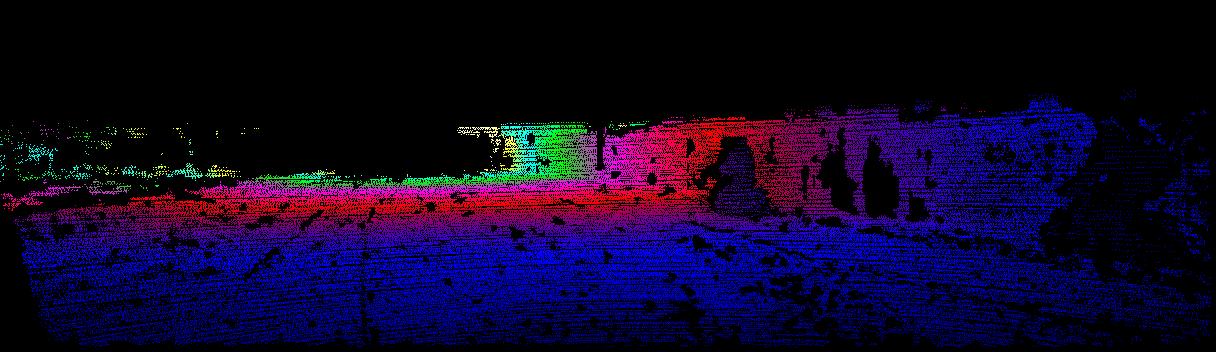} \\
  (c) semi-dense annotation
  \end{minipage}
  & 
  \begin{minipage}[m]{\figWidth}\centering
  \includegraphics[width=\linewidth]{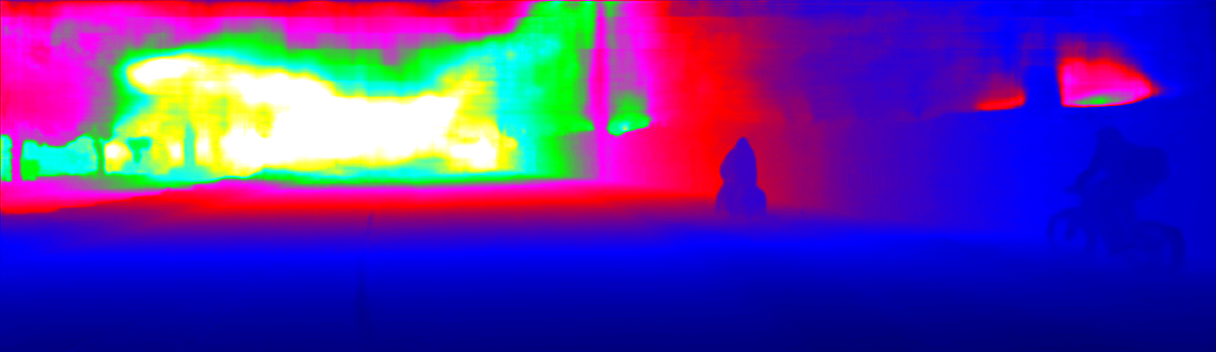} \\
  (d) dense prediction as depth image
  \end{minipage} \\
  \vspace{-1em} \\
  \multicolumn{2}{c}{
  \begin{minipage}[m]{\linewidth}\centering
  \includegraphics[width=\linewidth]{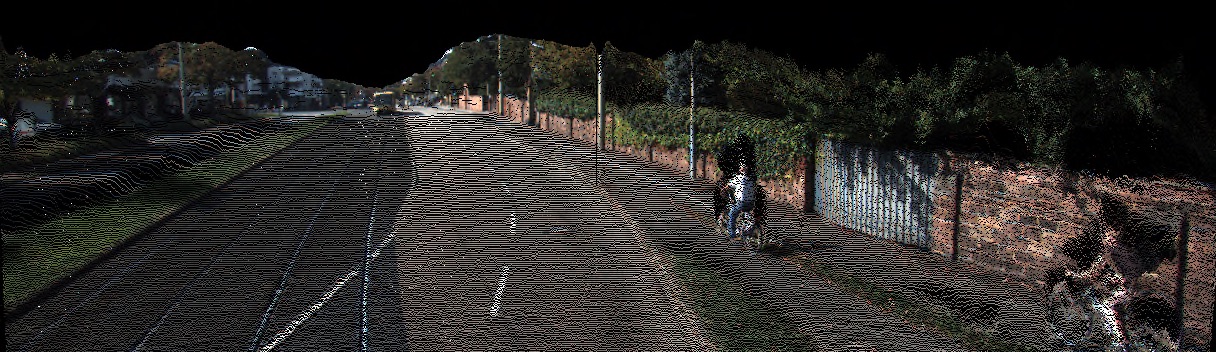} \\
  (e) dense prediction as point cloud
  \end{minipage}
  }
\end{tabular}

\end{minipage}

\caption{We develop a deep regressional network for \textit{depth completion}: given \textit{(a)} sparse \lidar scans, and possibly \textit{(b)} a color image, estimate \textit{(d)} a dense depth image. Semi-dense depth labels, illustrated in \textit{(d)} and \textit{(e)}, are generally hard to acquire, so we develop a highly-scalable, self-supervised framework for training such networks. Best viewed in color.}
\label{fig:RGBd}
\end{figure}
Depth sensing is fundamental in a variety of robotic tasks, including obstacle avoidance, 3D mapping~\cite{newcombe2011kinectfusion,zhang2014loam}, and localization~\cite{wolcott2015fast}. \lidar, given its high accuracy and long sensing range, has been integrated into a large number of robots and autonomous vehicles. 
However, existing 3D \lidars have a limited number of horizontal scan lines, and thus provide only sparse measurements, especially for distant objects (\eg, the 64-line Velodyne scan in \prettyref{fig:RGBd}~(a)). Furthermore, increasing the density of 3D \lidars measurements is cost prohibitive\footnote{Currently, the 16- and 64-line Velodyne \lidars cost around \$4k and \$75k, respectively}. Consequently, estimating dense depth from sparse measurements (\ie, \textit{depth completion}) is valuable for both academic research and large-scale industrial deployment.

Depth completion from \lidar measurements is challenging for several reasons. Firstly, the \lidar measurements are highly sparse and also irregularly spaced in the image space. Secondly, it is a non-trivial task to improve prediction accuracy using the corresponding color image, if available, since depth and color are different sensor modalities. Thirdly, dense ground truth depth is generally not available, and obtaining pixel-level annotations can be both labor-intensive and non-scalable.  

In this work, we address all these challenges with two contributions: 
\textit{(1)} We develop a network architecture that is able to learn a direct mapping from the sparse depth (and color images, if available) to dense depth. This architecture achieves state-of-the-art accuracy on the KITTI Depth Completion Benchmark~\cite{uhrig2017sparsity} and is currently the leading method.
\textit{(2)} We propose a self-supervised framework for training depth completion networks. Our framework assumes a simple sensor setup with a sparse 3D \lidar and a monocular color camera. The self-supervised framework trains a network without the need for dense labels, and outperforms some existing methods that are trained with semi-dense annotations. Our software\footnote{\url{https://github.com/fangchangma/self-supervised-depth-completion}} and demonstration video\footnote{\url{https://youtu.be/bGXfvF261pc}} will be made publicly available.


\section{Related Work}

\paragraph{Depth completion.}
\textit{Depth completion} is an umbrella term that covers a collection of related problems with a variety of different input modalities (\eg, relatively dense depth input~\cite{camplani2012efficient,shen2013layer,lu2014depth} vs. sparse depth measurements~\cite{ma2016sparse,ma2017sparse}; with color images for guidance~\cite{shen2013layer,ma2017sparse-to-dense} vs. without~\cite{uhrig2017sparsity}). The problems and solutions are usually sensor-dependent, and as a result they face vastly different levels of algorithmic challenges. 

For instance, depth completion for structured light sensor (\eg, Microsoft Kinect)~\cite{zhang2018deep} is sometimes also referred to as \textit{depth inpainting}~\cite{barron2016fast}, or \textit{depth enhancement}~\cite{camplani2012efficient,shen2013layer,lu2014depth} when noise is taken into account. The task is to fill in small missing holes in the relatively dense depth images. This problem is relatively easy, since most pixels (typically over 80\%) are observed. Consequently, even simple filtering-based methods~\cite{camplani2012efficient} can provide good results. 
As a side note, the inpainting problem also finds close connection to \textit{depth denoising}~\cite{diebel2006application} and \textit{depth super-resolution}~\cite{hornacek2013depth,xie2014single,lu2015sparse,xie2016edge,schneider2016semantically,jampani2016learning}.

However, the completion problem becomes much more challenging when the input depth image has much lower density, because the inverse problem is ill-posed. For instance, \citet{ma2016sparse,ma2017sparse} addressed depth reconstruction from only hundreds of depth measurements, by assuming a strong \textit{a priori} of piecewise linearity in depth signals. Another example is autonomous driving with 3D \lidars, where the projected depth measurements on the camera image space account for roughly 4\% pixels~\cite{uhrig2017sparsity}. This problem has attracted a significant amount of recent interest. Specifically, 
\citet{ma2017sparse-to-dense} proposed an end-to-end deep regression model for depth completion.
\citet{ku2018defense} developed a simple and fast interpolation-based algorithm that runs on CPUs. 
\citet{uhrig2017sparsity} proposed \textit{sparse convolution}, a variant of regular convolution operations with input normalizations, to address data sparsity in neural networks. \citet{eldesokey2018propagating} improved the normalized convolution for confidence propagation. 
\citet{chodosh2018deep} incorporated the traditional dictionary learning with deep learning into a single framework for depth completion. Compared with all these prior work, our method achieves significantly higher accuracy.

\paragraph{Depth prediction.}
Depth completion is closely related to depth prediction from a monocular color image. Research in depth prediction dates further back to early work by \citet{saxena2006learning}. Since then, depth prediction has evolved from simple handcrafted feature representations~\cite{saxena2006learning} to 
the deep learning based approaches~\cite{eigen2014depth,laina2016deeper,ummenhofer2016demon,fu2018deep} (see the reference therein). Most learning-based work relied on pixel-level ground truth depth training. However, ground truth depth is generally not available and cannot be manually annotated. 
%
To address such difficulties, recent focus has shifted towards seeking other supervision signals for training. For instance, 
\citet{zhou2017unsupervised} developed an unsupervised learning framework for simultaneous estimation of depth and ego-motion from a monocular camera, using photometric loss as a supervision. However, the depth estimation is only up-to-scale. 
\citet{mahjourian2018unsupervised} improved the accuracy by using 3D geometric constraints, and \citet{yin2018geonet} extended the framework for optical flow estimation.
\citet{li2017undeepvo} recovered the absolute scale by using stereo image pairs. 
%
In contrast, in this work we propose the first self-supervised framework that is designed specifically for depth completion. We utilize the \RGBd sensor data and the well-studied, traditional model-based methods for pose estimation, in order to provide absolute-scale depth supervision.


\section{Network Architecture}
We formulate the depth completion problem as a deep regression learning problem. For ease of notation, we use \sd for sparse depth input (pixels without measured depth are set to zero), \RGB for color images (or grayscale images), and \pred for depth prediction.

\begin{figure}[htbp]
\centering
\includegraphics[width=\linewidth]{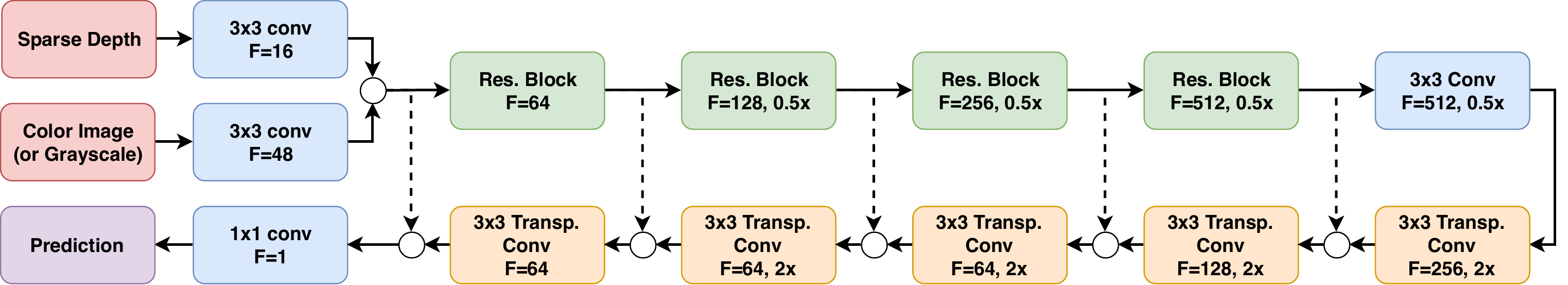}
\caption{Our deep regression network for depth completion, with both sparse depth and RGB as input. Skip connections are denoted by dashed lines and circles represent concatenation of channels.}
\label{fig:arch}
\end{figure}

The proposed network follows an encoder-decoder paradigm~\cite{ronneberger2015u}, as displayed in \prettyref{fig:arch}. The encoder consists of a sequence of convolutions with increasing filter banks to downsample the feature spatial resolutions. The decoder, on the other hand, has a reversed structure with transposed convolutions to upsample the spatial resolutions. 


The input sparse depth and the color image, when available, are separately processed by their initial convolutions. The convolved outputs are concatenated into a single tensor, which acts as input to the residual blocks of ResNet-34~\cite{he2016deep}. Output from each of the encoding layers is passed to, via skip connections, the corresponding decoding layers. A final 1x1 convolution filter produces a single prediction image with the same resolution as network input. All convolutions are followed by batch normalization~\cite{ioffe2015batch} and ReLU, with the exception at the last layer. 
At inference time, predictions below a user-defined threshold $\tau$ are clipped to $\tau$. We empirically set $\tau=0.9m$, the minimal valid sensing distance for \lidars. 

In the absence of color images, we simply remove the RGB branch and adopt a slightly different set of hyper parameters: the number of filters is reduced to half (\eg, the first residual block has 32 channels, instead of 64). 


\section{Self-supervised Training Framework}\label{sec:self-supervision}
Existing work on depth completion relies on densely annotated ground truth for training. However, dense ground truth generally does not exist, and even the acquisition of semi-dense labels can be technically challenging. For instance, \citet{uhrig2017sparsity} created an annotated depth dataset by aggregating consecutive data frames using GPS, stereo vision, and additional manual inspection. However, this method is not easily scalable. Furthermore, it produces only semi-dense annotations ($\sim30$\% pixels) within the bottom half of the image.

\begin{figure}[ht]
\centering
\includegraphics[width=\linewidth]{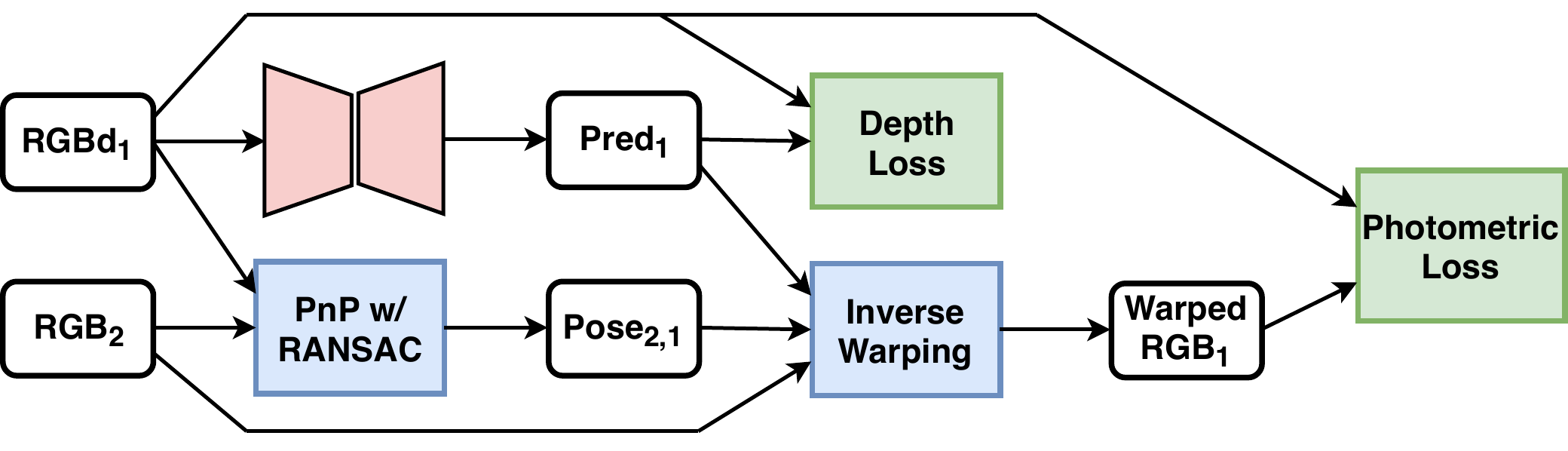}
\caption{An illustration of the self-supervised training framework, which requires only a sequence of color images and sparse depth images. White rectangles are variables, red is the depth network to be trained, blue are deterministic computational blocks (without learnable parameters), and green are loss functions.}
\label{fig:graph}
\end{figure}

In this section, we propose a model-based self-supervised training framework for depth completion. This framework requires only a synchronized sequence of color/intensity images from a monocular camera and sparse depth images from \lidar. Consequently, the self-supervised framework does not rely on any additional sensors, manual labeling work, or other learning-based algorithms as building blocks. Furthermore, this framework does not depend on any particular choice of neural network architectures.
The self-supervised framework is illustrated in \prettyref{fig:graph}. During training, the current data frame $\RGBd_1$ and a nearby data frame $\RGB_2$ are both used to provide supervision signals. However, at inference time, only the current frame $\RGBd_1$ is needed as input to produce a depth prediction $\pred_1$.

\paragraph{Sparse Depth Supervision}
The sparse depth input $\sd_1$ itself can be used as a supervision signal. Specifically, we penalize the differences between network input and output on the set of pixels with known sparse depth, and thus encouraging an identity mapping on this set. This loss leads to higher accuracy, improved stability and faster convergence for training. The depth loss is defined as
\begin{align}
\ldepth(\pred, \sd) = \norm{\indicator{\sd>0} \cdot (\pred - \sd)}_2^2.
\label{eqn:depth-loss}
\end{align}
Note that a denser ground truth (\eg, the 30\% dense annotation from the KITTI depth completion benchmark~\cite{uhrig2017sparsity}), if available, can also be used in place of the sparse input $\sd_1$. 

\paragraph{Model-based Pose Estimation}
As an intermediate step towards the photometric loss, the relative pose between the current frame and the nearby frame needs to be computed. Prior work assumes either known transformations (\eg, stereo~\cite{li2017undeepvo}) or the use of another learned neural network for pose estimation (\eg, \cite{zhou2017unsupervised}). In contrast, in this framework, we adopt a model-based approach for pose estimation, utilizing both \RGB and \sd. 

Specifically, we solve the Perspective-n-Point (PnP) problem~\cite{lepetit2009epnp} to estimate the relative transformation $T_{1\to2}$ between the current frame 1 and the nearby frame 2, using matched feature correspondences extracted from $\RGBd_1$ and $\RGB_2$ respectively. Random sample consensus (RANSAC)~\cite{fischler1987random} is also adopted in conjunction with PnP to improve robustness to outliers in feature matching. 
Compared to \RGB-based estimation~\cite{zhou2017unsupervised} which is up-to-scale, our estimation is scale-accurate and failure-aware (flag returned if no estimation is found). 

\paragraph{Photometric Loss as Depth Supervision}
Given the relative transformation $T_{1\to2}$ and the current depth prediction $\pred_1$, the nearby color image $\RGB_2$ can be inversely warped to the current frame. Specifically, given the camera intrinsic matrix $K$, any pixel $p_1$ in the current frame 1 has the corresponding projection in frame 2 as $p_2 = KT_{1\to2} \pred_1(p_1) K^{-1} p_1$. Consequently, we can create a synthetic color image using bilinear interpolation around the 4 immediate neighbors of $p_2$. In other words, for all pixels $p_1$:
\begin{align}
\texttt{warped}_1(p_1) = \textrm{bilinear}(\RGB_2(KT_{1\to2} \pred_1(p_1) K^{-1} p_1)).
\end{align}
\texttt{warped} is similar to the current $\RGB_1$ when the environment is static and there's limited occlusion due to change of view point. Note that this photometric loss is made differentiable by the bilinear interpolation. Minimizing the photometric error reduces the depth prediction error, only when the depth prediction is close enough to the ground truth (\ie, when the projected point $p_2$ differs from the true correspondence by no more than 1 pixel). Therefore, a multi-scale strategy is applied to ensure $\norm{p_2^{(s)}-p_1^{(s)}}_1<1$ on at least one scale $s$. In additional, to avoid conflicts with the depth loss, the photometric loss is evaluated only on pixels without direct depth supervision. The final photometric loss is
\begin{align}
\lphoto(\texttt{warped}_1, \RGB_2) = \sum_{s \in S} \frac 1 s \norm{\indicator{\sd==0}^{(s)} \cdot (\texttt{warped}_1^{(s)} - \RGB_2^{(s)})}_1,
\end{align}
where $S$ is the set of all scaling factors, and $(\cdot)^{(s)}$ represents image resizing (with average pooling) by a factor of $s$. Losses at lower resolutions are weighted down by $s$.

\paragraph{Smoothness Loss}
The photometric loss only measures the sum of all individual errors (\ie, color differences computed on each pixel independently) without any neighboring constraints. Consequently, minimizing the photometric loss alone usually results in an undesirable local optimum, where the depth pixels have incorrect values (despite having a low photometric error) and high discontinuity. To alleviate this issue, we add a third term to the loss functions in order to encourage smoothness of the depth predictions. Inspired by \cite{ma2017sparse,ma2016sparse,zhou2017unsupervised}, we penalize $\norm{\nabla^2 \pred_1}_1$, the $\mathcal{L}_1$ loss of the second-order derivatives of the depth predictions, to encourage piecewise-linear depth signal.

In summary, the final loss function for the entire self-supervised framework consists of 3 terms:
\begin{align}
\lself = \ldepth(\pred_1, \sd_1) + \beta_1~ \lphoto\left(\texttt{warped}_1, \RGB_1\right) + \beta_2~\norm{\nabla^2 \pred_1}_1
\end{align}
where $\beta_1, \beta_2$ are relative weightings. Empirically we set $\beta_1=0.1$ and $\beta_2=0.1$.


\section{Implementation}

For the sake of benchmarking against state-of-the-art methods, we use the KITTI depth completion dataset~\cite{uhrig2017sparsity} for both training and testing. The dataset is created by aggregating \lidar scans from 11 consecutive frames into one, producing a semi-dense ground truth with roughly 30\% annotated pixels. The dataset consists of 85,898 training data, 1,000 selected validation data, and 1,000 test data without ground truth.

For the PnP pose estimation, we dialate the sparse depth images $\sd_1$ with a $4 \times 4$ kernel, since the extracted features points might not have spot-on depth measurements. In each epoch, we iterate through the entire training dataset for the current frame 1, and choose a neighbor frame 2 randomly from the 6 nearest frames in time (excluding the current frame itself). In presence of PnP pose estimation failure, $T_{1\to2}$ is set to be an identity matrix and the neighbor $\RGB_2$ image is overwritten by the current $\RGB_1$. Consequently, the photometric loss is made to be 0, and does not affect the training.

The training framework is implemented in PyTorch~\cite{paszke2017automatic}. 
Zero-mean Gaussian random initialization is used for the network weights. We use a batch size of 8 for the \RGBd-network, and 16 for the simpler \sd-network. Adam with a starting learning rate of $10^{-5}$ is used for network optimization. The learning rate is reduced to half every 5 epochs. We use 8 Tesla V100 GPUs with 16G of RAM for training, and 12 epochs takes roughly 12 hours for the \RGBd-network and 4 hours for the \sd-network.


\section{Results}
In this section, we present experimental results to demonstrate the performance of our approach. We first compare our network architecture, trained in a purely supervised fashion, against state-of-the-art published methods. Secondly, we conduct an ablation study on the proposed network architecture to gain insight into which components contribute to the prediction accuracy. Lastly, we showcase training results using our self-supervised framework, and present an empirical study on how the algorithm performs under different level of sparsity in the input depth signals.


\subsection{Comparison with State-of-the-art Methods}
In this section, we train our best network in a purely supervised fashion to benchmark against other published results. We use the official error metrics for the KITTI depth completion benchmark~\cite{uhrig2017sparsity}, including \rmse, \mae, \irmse, and \imae. Specifically, \rmse and \mae stand for the root-mean-square error and the mean absolute error, respectively; \irmse and \imae stand for the root-mean-square error and the mean absolute error in the inverse depth representation. The results are listed in \prettyref{tab:sota} and visualized in \prettyref{fig:sota}.

\begin{table*}[ht]
\centering
\footnotesize
\setlength\tabcolsep{3pt} 
\caption{Comparison against state-of-the-art algorithms on the test set.
}
\label{tab:sota}
\begin{tabular}{| c || c || *{4}{ c |} }
\hline
Method & Input & \rmse~[mm] & \mae~[mm] & \irmse~[1/km] & \imae~[1/km]\\ \hline \hline
NadarayaW~\cite{uhrig2017sparsity} & \sd & 1852.60 & 416.77 & 6.34 & 1.84 \\
SparseConvs~\cite{uhrig2017sparsity} & \sd & 1601.33 & 481.27 & 4.94 & 1.78 \\
ADNN~\cite{chodosh2018deep} & \sd & 1325.37 & 439.48 & 59.39 & 3.19 \\
IP-Basic~\cite{ku2018defense} & \sd & 1288.46 & 302.60 & 3.78 & \textbf{1.29} \\
NConv-CNN~\cite{eldesokey2018propagating} & \sd & 1268.22 & 360.28 & 4.67 & 1.52 \\
NN+CNN2~\cite{uhrig2017sparsity} & \sd & 1208.87 & 317.76 & 12.80 & 1.43 \\
Ours-\sd & \sd & \textbf{954.36} & \textbf{288.64} & \textbf{3.21} & 1.35 \\
\hline
SGDU~\cite{schneider2016semantically} & \RGBd & 2312.57 & 605.47 & 7.38 & 2.05 \\
Ours-\RGBd & \RGBd & \textbf{814.73} & \textbf{249.95} & \textbf{2.80} & \textbf{1.21} \\
\hline
\end{tabular}
\end{table*}

Our \sd-network leads prior work with a large margin in almost all metrics. The \RGBd-network attains even higher accuracy, leading all submissions to the benchmark. Our predicted depth images also have cleaner and sharper object boundaries (\eg, see trees, cars and road signs), which can be attributed to the fact that our network is quite deep (and thus might be able to learn more complex semantic representations) and has large skip connections (and thus preserves image details). Note that all these supervised methods produce poor predictions at the top of the image, because of 2 reasons: (a) the \lidar returns no measurements, and thus the input to the network is all zero at the top; (b) the 30\% semi-dense annotations do not contain labels in these top regions.

\begin{figure*}[ht]
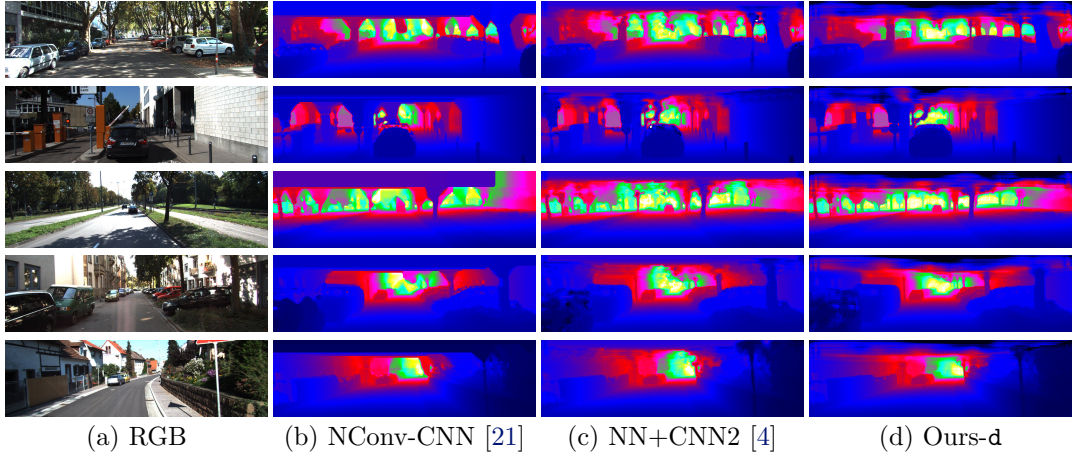

\centering
\newcommand{\loadfig}[1]{\includegraphics[width=0.25\linewidth]{{figures/test/#1}}}
\newcommand{\createrow}[1]{\loadfig{000000000#1_rgb} & \loadfig{000000000#1_nconv_cnn} & \loadfig{000000000#1_nn_cnn2} & \loadfig{000000000#1_s2d2_d}}
\setlength\tabcolsep{1pt} 
\begin{tabular}{*{4}{ c }}
\createrow{0} \\
\createrow{1} \\
\createrow{2} \\
\createrow{3} \\
\createrow{4} \\
(a) RGB & (b) NConv-CNN~\cite{eldesokey2018propagating} & (c) NN+CNN2~\cite{uhrig2017sparsity} & (d) Ours-\sd\\
\end{tabular}
\caption{Comparision against other methods (best viewed in color). Our predictions have not only lower errors, but also cleaner and sharper boundaries.}
\label{fig:sota}
\end{figure*}


\subsection{Ablation Studies}\label{sec:ablation}
To examine the impact of network components on performance, we conduct a systematic ablation study and list the results column-wise in \prettyref{tab:ablation}. 


\begin{table*}[ht]
\newcommand{\filter}[1]{$F_1\!=\!#1$\xspace}
\centering
\small
\setlength\tabcolsep{2pt} 
\caption{Ablation study of the network architecture for depth input. Empty cells indicate the same value as the first row of each section. See \prettyref{sec:ablation} for detailed discussion.}
\label{tab:ablation}
\vspace{0.5em}
\begin{tabular}{|  *{10}{ c } || c |}
\hline
image & \multilinecell{fusion\\split} & loss & \multilinecell{ResNet\\depth} & \multilinecell{with\\skip} & \multilinecell{reduced\\filters} & \multilinecell{pre-\\trained} & \multilinecell{\textnumero \\ pairs} & \multilinecell{down-\\sample} &\multilinecell{dropout\\ \& weight\\decay} & \multilinecell{\rmse\\~[mm]}  \\

\hline\hline
None & - & $L_2$ &34 & Yes & 2x (\filter{32}) &  No&5 & No & No & \textbf{991.35} \\

\hline
&&$L_1$&&&&&&&&1170.58\\

\hline
&&&18&&&&&&&1003.78\\

\hline
&&&&No&&&&&&1060.64\\

\hline
&&&&&1x (\filter{64})&&&&&992.663\\

\hline
&&&&&1x (\filter{64})&Yes&&&&1058.218\\

\hline
&&&&&4x (\filter{16})&&&&&1015.204\\

\hline
&&&&&&&4&&&996.024\\

\hline
&&&&&&&3&&&1005.935\\

\hline
&&&&&&&&Yes&&1045.062\\

\hline
&&&&&&&&&Yes&1002.431\\

\hline \hline
Gray & 16/48 &$L_2$ &  34 & Yes & 1x (\filter{64}) &  No& 5 & No & Yes & \textbf{856.754} \\

\hline
RGB&&&&&&&&&&859.528\\

\hline
&32/32&&&&&&&&&868.969\\


\hline
&&&18&&&&&&&875.477\\

\hline
&&&&No&&&&&&1070.789\\

\hline
&8/24&&&&2x (\filter{32})&&&&&887.472\\


\hline
&&&&&&&4&&&857.154\\

\hline
&&&&&&&3&&&857.448\\

\hline
&&&&&&&&Yes&&859.528\\

\hline
\end{tabular}
\end{table*}

The most effective components in improving final accuracy includes using \RGBd for input and $\ltwo$ loss for training. This is in contrary to the findings that $\lone$ is more effective \cite{ma2017sparse-to-dense,carvalhoregression}, implying that the optimal loss functions might be dataset- and architecture-dependent.
Adding skip connections, training from scratch (without ImageNet-pretraining), and not using max pooling also result in substantial improvement. Increasing network depth (from 18 to 34) and encoders-decoders pairs (from 3 to 5), as well as a proper split of filters allocated to the \RGB and the \sd branches (16/48 split), also create small positive impact on the results. However, additional regularization, including dropout combined with a weight decay, leads to degraded performance. 

It is worth noting that alternative encoding of the input depth image (such as the nearest neighbor interpolation or the bilinear interpolation of the sparse depth measurements) does not improve the prediction accuracy. This implies that the proposed network is able to deal with highly sparse input image. 

\subsection{Evaluation of the Self-supervised Framework}\label{sec:exp-self}
In this section, we evaluate the self-supervised training framework described in \prettyref{sec:self-supervision} on the KITTI validation dataset. We compare 3 different training methods: using only photometric loss without sparse depth supervision, the complete self-supervised framework (\ie, photometric loss with sparse depth supervision), and the pure supervised method using the semi-dense annotations. The quantitative results are listed in \prettyref{tab:self-supervision}. The self-supervised result produces $\text{\rmse}=1384$, which already outperforms some of the prior methods that were trained with semi-dense annotations, such as SparseConvs~\cite{uhrig2017sparsity}. 

\begin{table*}[ht]
\centering
\footnotesize
\setlength\tabcolsep{4pt} 
\caption{Evaluation of the self-supervised framework on the validation set
}
\label{tab:self-supervision}
\begin{tabular}{| c || *{4}{ c |}  }
\hline
Training Method & \rmse~[mm] & \mae~[mm] & \irmse~[1/km] & \imae~[1/km]\\ \hline \hline
Photometric Loss Only & 1901.16 & 658.13 & 5.85 & 2.62 \\
Self-Supervised  & 1384.85 & 358.92 & 4.32 & 1.60 \\
Supervised Learning & 878.56 & 260.90 & 3.25 & 1.34 \\
\hline
\end{tabular}
\end{table*}

\begin{figure*}[ht]
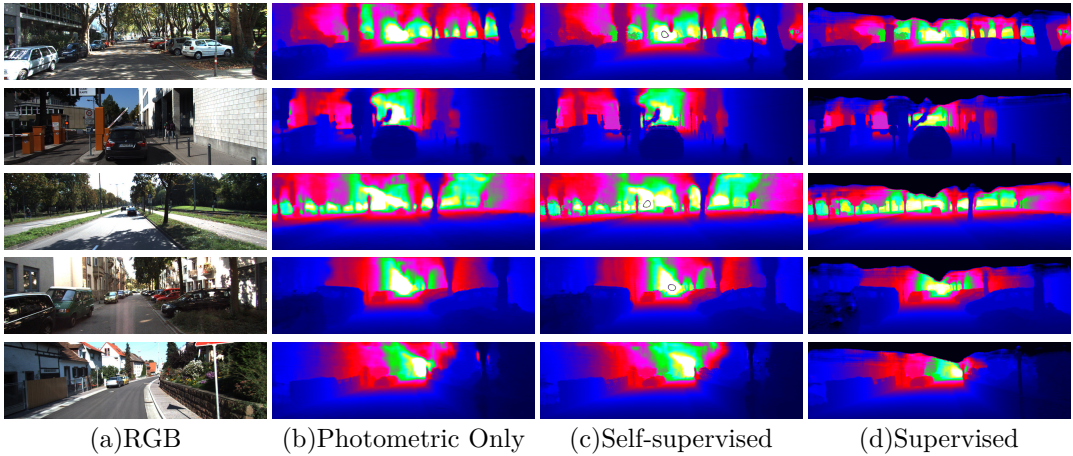

\centering
\newcommand{\loadrgb}[1]{\includegraphics[width=0.25\linewidth]{{figures/test/#1}}}
\newcommand{\loaddepth}[2]{\includegraphics[width=0.25\linewidth]{{figures/self/#1/#2.png}}}
\newcommand{\loadgray}{\includegraphics[width=0.125\linewidth]{{figures/gray.jpg}}}
\newcommand{\createrow}[1]{\loadrgb{000000000#1_rgb} & \loaddepth{photo_only}{out#1} & \loaddepth{photo_sparse}{out#1} & \loaddepth{dense}{out#1}}
\setlength\tabcolsep{1pt} 
\begin{tabular}{*{4}{ c }}
\createrow{0} \\
\createrow{1} \\
\createrow{2} \\
\createrow{3} \\
\createrow{4} \\
(a)RGB & (b)Photometric Only & (c)Self-supervised & (d)Supervised\\
\end{tabular}
\caption{Comparision between different training methods (best viewed in color). The photometric loss provides supervision at the top, where the semi-dense annotation does not contain labels.}
\label{fig:self-supervision}
\end{figure*}

However, note that the true quality of depth predictions trained in a self-supervised fashion is probably underestimated by such evaluation metrics, since the ``ground truth'' itself is biased. Specifically, the evaluation ground truth is characterized by the same limitations as the training annotations: low-density, as well as absence at the top region. As a result, predictions at the top, where the self-supervised framework provides supervision but semi-dense annotations do not, are not reflected in the error metrics, as illustrated in \prettyref{fig:self-supervision}.

The self-supervised framework is effective for not only 64-line lidar measurements, but also lower-resolution lidars and more sparse depth input. In \prettyref{fig:s2d}(b), we show the validation errors of the networks trained with the self-supervised framework with different levels of sparsity in the depth. When the number of input measurements is too small, the validation error is high. This is expected due to failure in PnP pose estimation. However, with sufficiently many measurements (\eg, at least 4 scanlines, or the equivalent number of samples to at least 2 scanlines when input is uniformaly sampled), the validation error starts to decrease as a power function of the input, similar to training with semi-dense annotations.


\subsection{On Input Sparsity}
\begin{figure}[ht]
\centering
\newcommand{\figWidth}{0.49\textwidth}
\begin{minipage}{\figWidth}
\centering
\includegraphics[width=\linewidth]{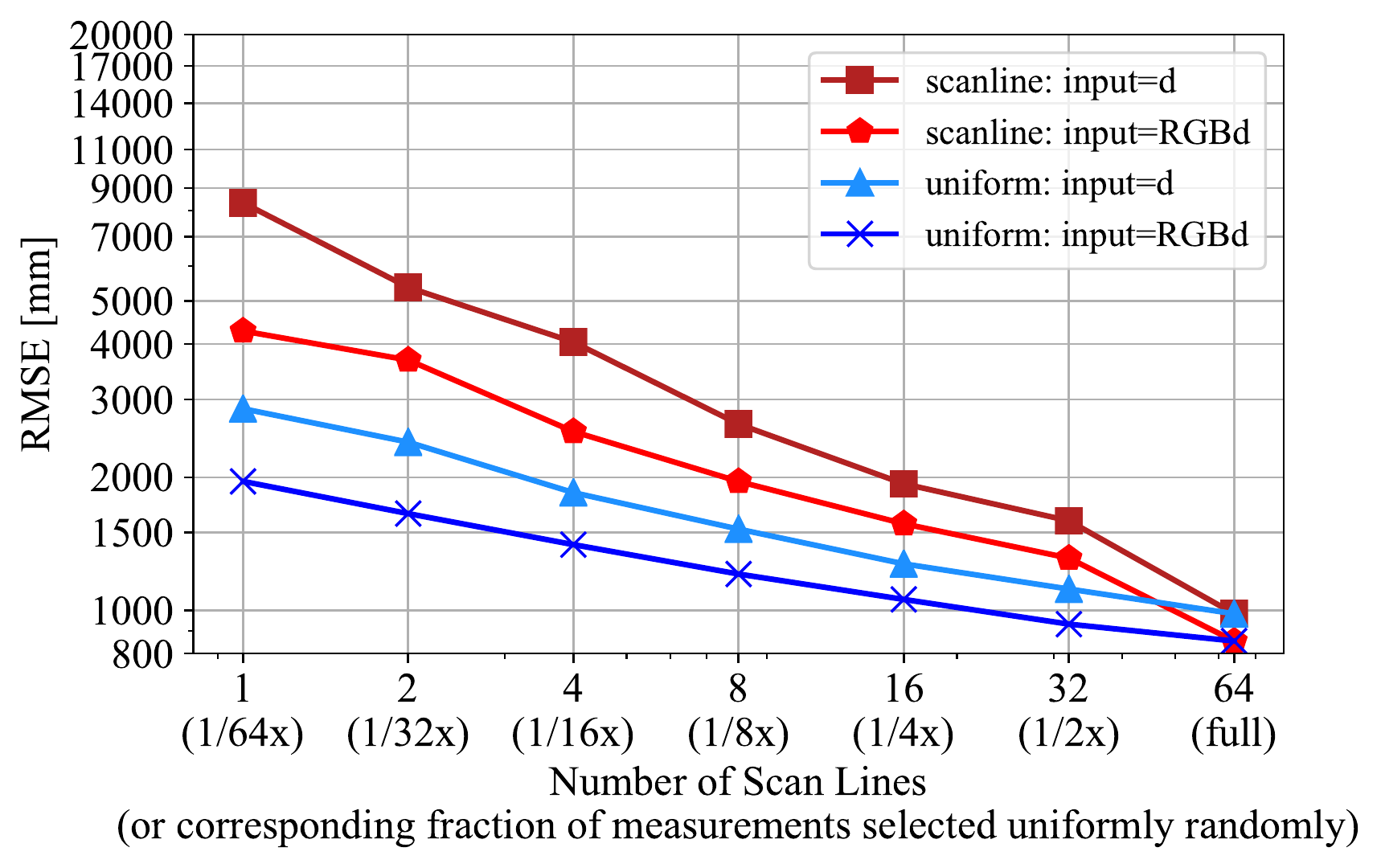}
\\(a) trained with semi-dense annotations
\end{minipage}
\begin{minipage}{\figWidth}
\centering
\includegraphics[width=\linewidth]{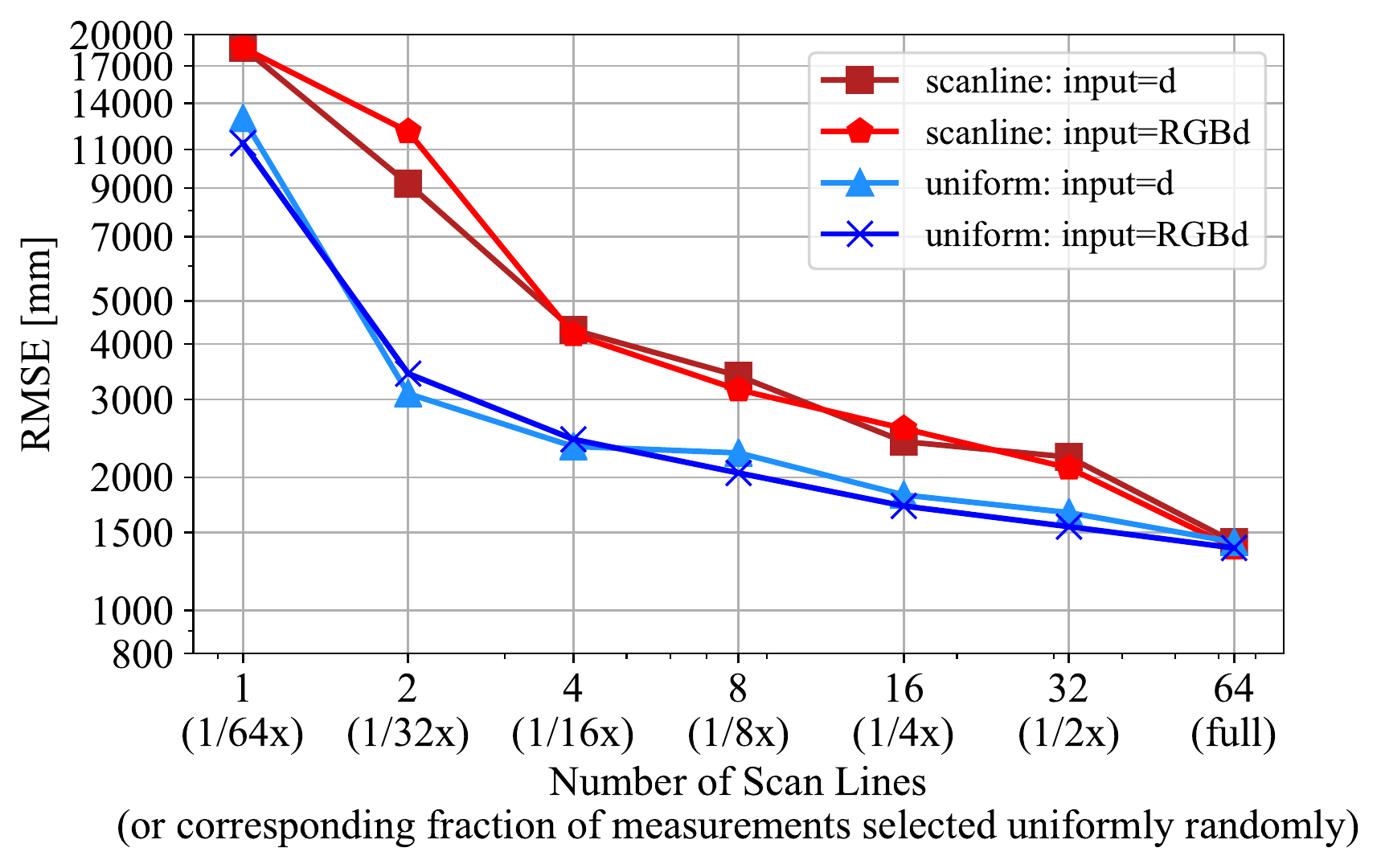}
\\(b) self-supervised
\end{minipage}
\caption{Prediction error against number of input depth samples, for both spatial patterns (uniform random sub-sampling and \lidar scan lines). (a) When trained with semi-dense ground truth, the depth completion error decreases as a power function $cx^p$ of the number of input depth measurements, for some $c>0, p<0$. (b) The self-supervised framework is effective with sufficiently many measurements (at least 4 scanlines, or the equivalent number of samples to 2 scanlines when input is uniformaly sampled).}
\label{fig:s2d}
\end{figure}

In many robotic applications, engineers need to address the following question: \textit{what's the \lidar resolution (which translates to financial cost) required to achieve certain performance?}
In this section, we try to answer this question by evaluating the accuracy of our \lidar depth completion technique under different input sparsity and spatial patterns. 
To this end, we provide an empirical analysis on the depth completion accuracy for different depth input with varying levels of sparsity and spatial patterns. In particular, we downsample the raw \lidar input in two different manners: reducing the number of laser scans (to simulate a \lidar with fewer scan lines), and uniformly sub-sampling from all \lidar measurements available. The results are illustrated in \prettyref{fig:s2d}, for both of these spatial patterns and both input modalities of d and RGBd.


In \prettyref{fig:s2d}(a) we show the validation errors when trained with semi-dense annotations. The \rmse errors form a straight line in the log-log plot, implying that the depth completion error decreases as a power function $cx^p$ of the number of input depth measurements, for some positive $c$ and negative $p$. 
This also implies diminishing returns on increasing \lidar resolutions. 
Comparing the two spatial patterns, uniform random sub-sampling produces significantly higher accuracy than having a reduced number of scan lines, since the input depth samples are more disperse in the pixel space with uniform random sampling. Furthermore, using \RGBd substantially reduces prediction error, compared to using only \sd, when trained with semi-dense annotations. The performance gap is especially significant when the number of depth measurements is low. Note that there is a significant drop of RMSE from 32-line to 64-line \lidar. This accuracy gain may be attributed to the fact that our network architecture is optimized for 64-line \lidar. 

In \prettyref{fig:s2d}(b), we show results when trained with our self-supervised framework. As has been discussed in \prettyref{sec:exp-self}, the validation error starts to decrease steadily as a power function, similar to training with semi-dense annotations, when there are sufficiently many input measurements. However, with the self-supervised framework, using both RGB and sparse depth yields the same level of accuracy as using sparse depth only, which is different from training with semi-dense annotations. The underlying cause of this difference remains to be further investigated\footnote{In the self-supervised framework, the training process is more iterative than training with semi-dense annotations. In particular, it takes many more iterations for the predictions to converge to the correct value.
Consequently, the network weights for the RGB input, which has substantially lower correlation with the depth prediction than the sparse depth input, might have dropped to negligible levels during early iterations, resulting in similar performance for using d and RGBd as input. However, this conjecture remains to be verified.}.



\section{Conclusions}\label{sec:conclusion}
In this paper, we have developed a deep regression model for depth completion of sparse \lidar measurements. Our model achieves state-of-the-art performance on the KITTI depth completion benchmark, and outperforms existing published work by a significant margin at the time of submission. We also propose a highly scalable, model-based self-supervised training framework for depth completion networks. This framework requires only sequences of RGB and sparse depth images, and outperforms a number of existing solutions trained with semi-dense annotations. Additionally, we present empirical results demonstrating that depth completion errors decrease as a power function with the number of input depth measurements. In the future, we will investigate techniques for improving the self-supervised framework, including better loss functions and taking dynamic objects into account.



\acknowledgments{This work was supported in part by the Office of Naval Research (ONR) grant N00014-17-1-2670 and the NVIDIA Corporation. In particular, we gratefully acknowledge the support of NVIDIA Corporation with the donation of the DGX-1 used for this research. Finally, we thank Jonas Uhrig and Nick Schneider for providing information on how data is generated for the KITTI dataset~\cite{uhrig2017sparsity}.}

\bibliography{references/depth_completion,references/depth_prediction,references/mapping}

\end{document}